# Recent Developments in the Optimization of Space Robotics for Perception in Planetary Exploration

The Optimization of Autonomous Extra Terrestrial Rovers

S. Ahsan Badruddin
Department of Aeronautics and Astronautics,
Institute of Space Technology,
Islamabad, Pakistan
sahsanb@gmail.com

S. M. Dildar Ali
Department of Aeronautics and Astronautics,
Institute of Space Technology,
Islamabad, Pakistan
dildar_jaffar@yahoo.com

*Abstract*—The following paper reviews recent developments in the field of optimization of space robotics. The extent of focus of this paper is on the perception (robotic sense of analyzing surroundings) in space robots in the exploration of extra-terrestrial planets. Robots play a crucial role in exploring extra-terrestrial and planetary bodies. Their advantages are far from being counted on finger tips. With the advent of autonomous robots in the field of robotics, the role for space exploration has further hustled up. Optimization of such autonomous robots has turned into a necessity of the hour. Optimized robots tend to have a superior role in space exploration. With so many considerations to monitor, an optimized solution will nevertheless help a planetary rover perform better under tight circumstances. Keeping in view the above mentioned area, the paper describes recent developments in the optimization of autonomous extra-terrestrial rovers.

## I. Introduction

Over last few decades, the application of robots has increased drastically not only in industrial sector, but also in many engineering and technological applications [1]. One of the key most lead of today's robots over humans is that these machines can perform various tasks efficiently round the clock and without a fatigue in most rugged environments where mankind could not survive or unable to perform desired job. For example, exploring inside the volcano, and under water search missions [2].

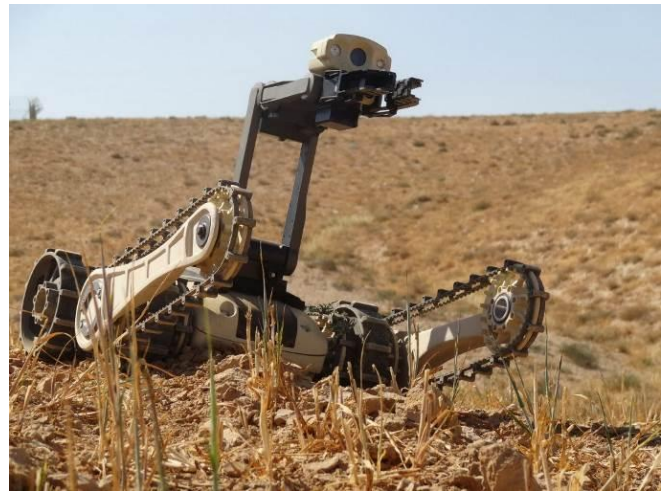

**Fig. 1 - A Micro Tactical Ground Robot (MTGR) being tested on harsh terrain [3]**

Robots facilitate exploring operations in deep sea, on planetary surfaces [4], and in orbits. It has a vital role in discovering new worlds of space, as well as to perform certain operations where mankind cannot survive [5]. At present, the field of robotics has expanded so enormous that it has crossed the confines of this world by entering into upper atmosphere and space, introducing the new category of robots, the 'Space Robots' [6]. After the successful robotic missions on Moon and Mars, future plans have been made for carrying out various activities and experiments [8] by staying longer period of time in upper atmosphere. Space missions require massive presence of robots, where they can be assigned to perform crucial operations without any human supervision [2].

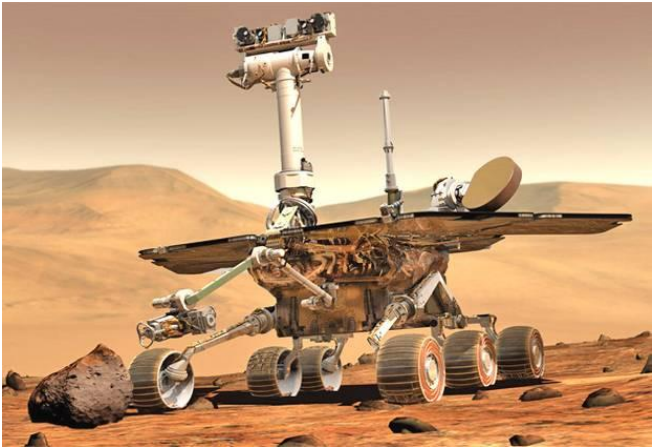

**Fig. 2 - Mars Rover [7]**

Exploration demands and gives freedom to space mission rovers. Among different exploration techniques, the best technique is the one which provides a mirror of the map or nearly an approximation of the map in a shorter period [9]. Generally, a rover robot demands non-deterministic action request because of unpredictable terrain where it operates [10]. The critical task of autonomous space robot or a rover is to move forward and intelligently explore the space without being frequently contacted by Earth station or mission operators [11]. The success of these rovers in any planetary surface mission strongly depends on the robot's ability to sense, identify, and perceive information about its unstructured surroundings and unexplored environment [12]. The significant-most concern for any autonomous rover is to explore paths in an unknown environment [13]. It is nearly impossible for a robot to be controlled remotely because of thousands of kilometers of distance and propagation delays of signals associated with atmospheric conditions. Therefore, the robot has to adapt to the poorly identified working environment [14].

Among robotic skills, it is obvious that the perception capability (so as vision in turn [16]) is the most significant feature in all complex and autonomously performed operations [2]. For an entirely autonomous rover, an effective way for learning from demonstration should be the ability to reveal descriptions of tasks and skills contained in a demonstration database [17]. For an outdoor application of mobile robots, multiple sensors are commonly used [18] to evaluate and measure the range and presence of obstacles, robot direction, location in the environment, and robot motions.

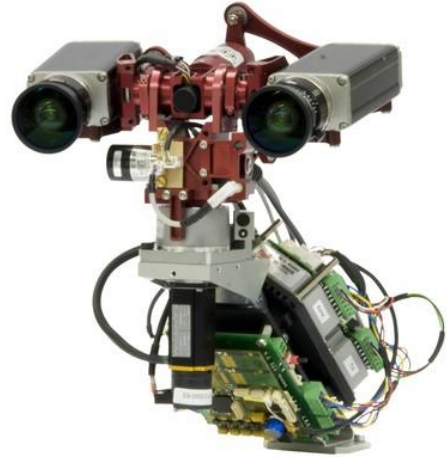

**Fig. 3 - An autonomous robot with perception capability [15]**

Different types of sensors are used for different physical phenomena and operations (e.g. optical, inertial, and magnetic) [12]. In the absence of an appropriate and efficient sensing system, the robot will not be able to perform required tasks and will not be able to handle the unexpected environmental obstructions [19].

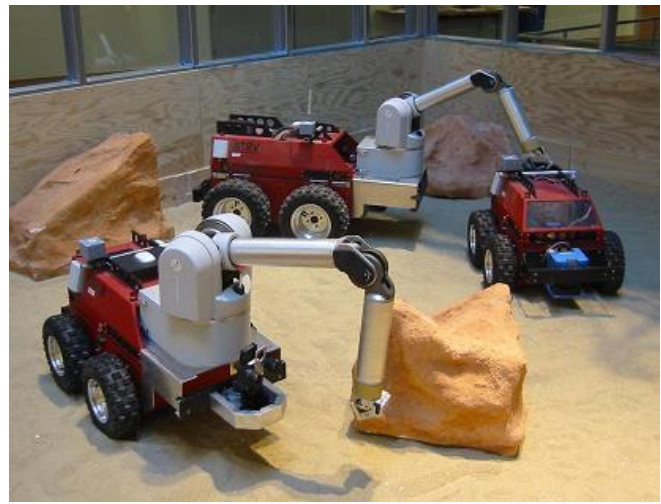

**Fig. 4 - A CSAIL robust robotic system [20]**

The future demands of robust robotic systems for space exploration missions will continue to increase at much higher levels in the near future [5]. Space robots are relatively inaccessible with respect to impact on mission of necessary communication with earth based station which further provokes to fully implement the autonomous capabilities [21]. For the next decade, it will be sufficient for carrying out limited missions through the traditional approach of using few specialized robots varying in capabilities to cover the complete span of mission requirements. For a permanent presence of human and robotic outposts in upper space (on Moon and Mars), the need for more efficient and fully autonomous robotic systems will increase, undoubtedly [22, 23].

## II. THE NEED FOR PERCEPTION CAPABILITIES

The design of autonomous rovers [24] should be amply smart, such that the rover can navigate in an unknown environment with obstacles (rocks, rough tracks & boulders) even in hazardous situations. While moving and navigating on planetary surfaces, the rover may encounter steep inclinations, sand covered pits, cliffs, ditches, and other uneven hindrances. The robot must avoid itself from surface hazards through intelligent negotiations from obstacle to obstacle, without any supervision, in order to achieve the assigned scientific exploration objectives in the natural environment [12]. After the latest Mars Exploration Rover (MER) mission, it was observed that important developments in robot's autonomy and navigation are still desirable [25].

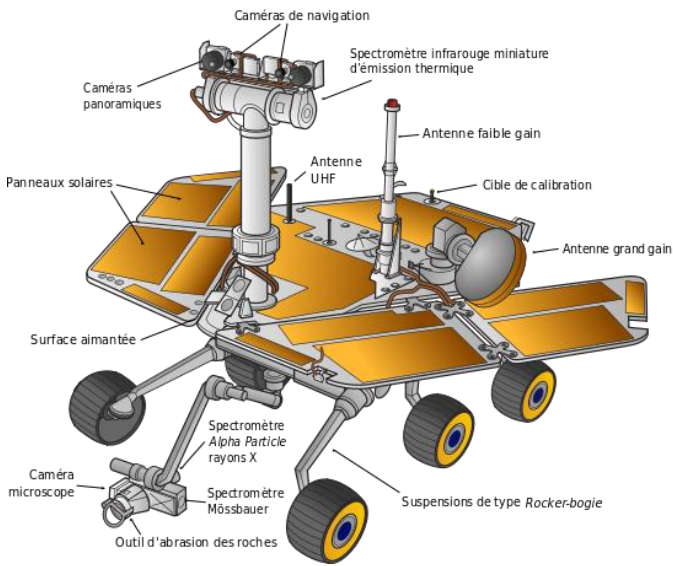

**Fig. 5 - The MER vehicle [26]**

The rover seeks for the best possible path in accordance with the mission objectives [27]. Entire movements of the rover are supplements of the description of its surroundings or environment, the combination of actions performed in that description and thus obtains the resulting situation to make the practice of the system [28].

To move independently in such extreme environments, a rover must identify and estimate the mobility risk and then discover the right path in the right direction. Moreover, several problems for navigation include maintaining prior information about rover's position, direction, and attitude on terrain. Also, the concern for mapping local surroundings and prominent landmarks should not be left unattended [29]. For these reasons, sensing and perception capabilities for successful navigation, are crucial for space missions like land reconnaissance and survey of topography for scientific exploration purposes [12].

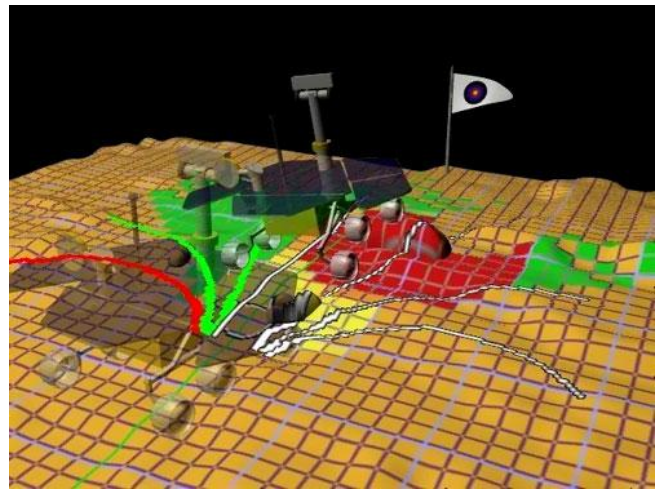

**Fig. 6 - Rover path finding [30]**

Generally, rover path finding, motion planning and controls are first simulated before getting applied to a real system [31]. In order to complete multiple real-world tasks in unpredictable environments [32, 33], it is indispensable to simultaneously join mapping and localization, path finding and planning, obstacle dodging and waypoint behaviors [33] with the physical dexterity. Visual perception and autonomous manipulation capabilities are also necessary equivalents for human hands [33, 34]. There is a huge need for intelligent strategies for robots to accomplish their assigned activities by avoiding collision or crash with obstacles and with one and other (in case of multiple rovers) [35]. Also, they should manage themselves in overcoming unknown obstacles in the rough environment [36].

## III. CONSTRAINTS

Regardless of the wide availability and variety of sensor technologies, space robot sensor systems cannot take full benefits of themselves because of limitations and constraints concerning mass, size (volume), power and survivability in space environment [12]. Rugged and complex terrains can then be in the capacity for designs of such robots [37]. The vision algorithm performance also has constraints and strongly depends on satellite models and features of natural images [38].

**Table 1 - Human sense and equivalent robotic sensors [39]**

| Human Sense | Robotic Sensor |
|---|---|
| Vision | Camera, Infrared Camera, Radiation Sensors |
| Audition | Microphones, Hydrophones |
| Gustation | Chemical Sensors |
| Oflaction | Chemical Sensors |
| Tactition | Contact Sensors, Force Sensors |
| Proprioception | Wheel Encoders |
| Equilibrioception | Tilt Sensors, Accelerometers, Gyroscope |
| Thermoception | Thermocouple |

Generally sensors or sensing systems used by rovers must fulfill certain specifications/criteria in accordance with rover payload capacity, size, onboard available power, thermal issues, and radiation tolerances. These checks are mainly imposed due to unique characteristics of targeted space environment and space mission objectives. Preferred solutions should be mechanically simple, have low mass, be low power operated and has to be airborne qualified [12, 40]. Similarly, sensor electronics should be tested and qualified for extreme thermal ranges so that rovers can operate in harsh temperature and extreme radiated environments of upper space as well as on thin atmosphere based planetary surfaces (such as Mars) [12]. It is also evident that mobile rovers can malfunction and become unreliable and unstable for various reasons like structural failure, mechanical failure, electronics breakdown, or computational crash [41]. One more critical issue lies with the accuracy and precision of position sensors during longer runs [42].

Constraints of sensors and sensor systems are treated as hard constraints as reliability of outer space tasks on hardware is very critical. The reason is that no hardware repair and maintenance can be performed once the rover is launched into space. For every mission, post-launch activities should be carried out without any failure or malfunctioning of hardware system. That is why massive research is performed for the selection of any particular design or configuration of sensor electronics as well as mechanical assemblies. For instance, in hazard detection sensing system, solid state solutions are preferred instead of mechanically scanning system [12, 40].

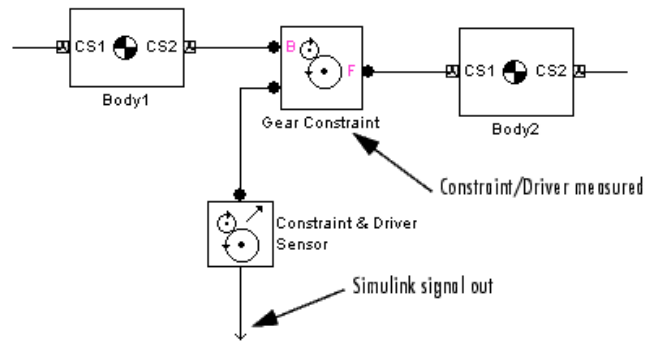

**Fig. 7 - Block diagram for sensor constraints**

Development of smart and intelligent mediators sets imperative scientific queries [43] for the field of robotics and artificial intelligence [44]. Moreover, for mechanically actuated sensor systems, there is a need of no or less moving parts configuration. Such designs have higher probability to withstand vibrations, gravity forces on space launched vehicle, and landing impacts related to flights from the Earth to other planetary surfaces [12].

Controller design of autonomous mobile rovers often needs consideration of goals and specifications [46]. Recently, significant advancements have been made in the manufacturing field of robots and actuators [47]. To date, available technology and sensor options have made possible limited success in our ability to build intelligent autonomous robots or robotic vehicles, despite of major advances in computing technology and intelligent algorithms for autonomy. Indeed, algorithms and computations are only parts of any solution. From future perspective, there is an extreme need of improvement and novelty in sensing solutions to move on to the next level [12].

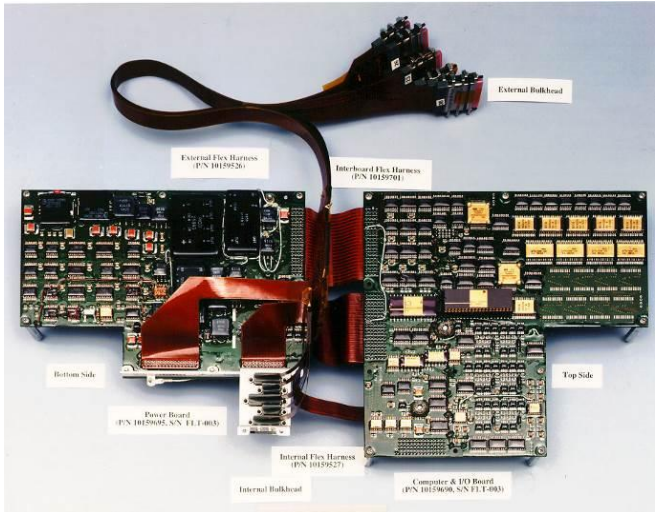

**Fig. 8 - Controller for the Mars Pathfinder [46]**

## IV. OPTIMIZATION

Certain action plans are needed in order to allow the robot to complete its tasks, without causing any disruption in the physical constraints such as saturation of actuator, consumption of energy, kinematic constraints, etc. Here is where optimization comes into play. These plans contain scripts on task instructions, sensing and navigation and are generated online from physically attainable actions; using a hierarchical process including genetic algorithm. The action plan thus enables the robots to be capable of performing the tasks in complex, rough and irregular environments [48] by preventing the system from being halted [49].

Several current action plans however do not take into consideration the physical characteristics of both the robot and the environment which limits the efficiency of the former. The inconsideration of physical characteristics along with limited human supervision would limit the robots in future to perform difficult tasks in rugged terrain [50]. Certain analysis and interpretation is performed on various sensory inputs (like camera feed, proximity sensors etc.) to build the robotic perception [51]. It should also be taken into account that the perceiving ability of rovers has certain amount of uncertainty [52] while moving on planetary surfaces. These aspects might be overlooked yet they are somehow obvious [53]. If it is desirable to permit a robot for further demonstrations, operator's acceptance and willingness should be increased for interaction with robot by means of sophisticated deigns [54].

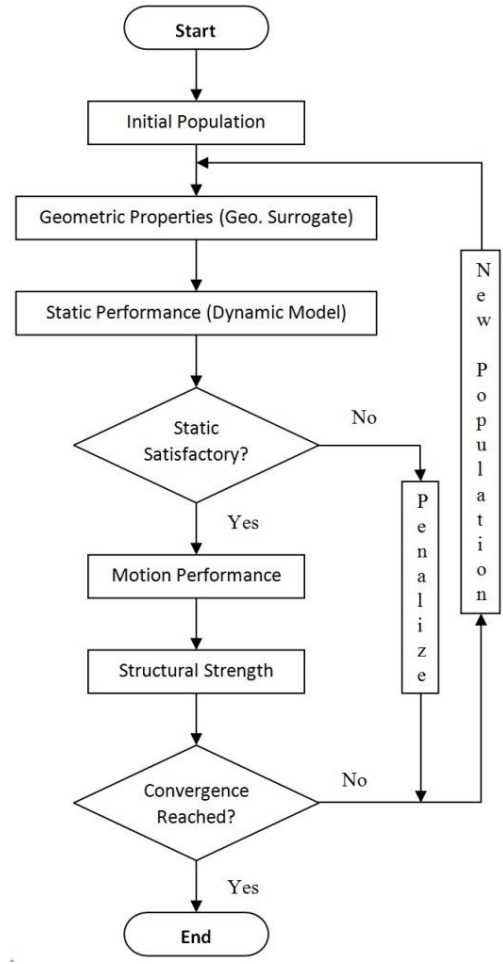

**Fig. 9 - Multidisciplinary Design Optimization of a Modular Robot**

### A. Task and Configuration Filters

The hierarchical selection process is based on the following two steps:

*a) Application of the module filters based on physical configuration considerations*

Simple tests are used by these filters based on the behavior of the problem. This step aims at removing certain components of the action plan that do not require the application of complex evaluation techniques such as detailed stimulation. This elimination helps in reduction of the number of possible plans [49].

*b) Application of configuration based module filters*

If the construction of the robot is based on modular components, certain modules may be included in the action module inventory, which are not concerned with the specific configuration [49].

## B. *Genetic Algorithm*

A genetic algorithm (GA) is a technique inspired by biological processes of natural selection and evolution that aid in finding the most appropriate action plan. The solutions on which genetic algorithms operate are called its generation. In order to perform its task, genetic algorithm works over the crossover and mutation operators and a fitness function [55, 56]. Genetic algorithm has been deduced from natural phenomena and works on the principle that only the strongest chromosome will survive [57]. Over the last decade, heuristic algorithms (especially GA) have extensively been implemented to generate the best possible pathway through utilization of its powerful optimization technique and methodology [58]. The notable strength of genetic algorithm is because of parallel and simultaneous search of the best solution in the entire search space [59], which is performed through a generation of solution's population [60].

The crossover operator is a process which mimics the biological process of crossover in which two chromosomes crossover, exchange their parts and result in a unique chromosome. In the similar manner, this process aims at exchanging a random module in the action plan and its following modules with a random module of a second action plan and its following modules, thus producing a new action plan [49]. The mutation operator is a process that is used to preserve the diversity in a population. In this process a module from one action plan is switched with a module opted from the reduced inventory [49, 55]. Fitness function is a method that helps determine the fitness of an action plan. In order to do that, a simulation is used to implement the plan to find out whether the robot accomplishes its tasks successfully or not. The simulation checks for factors such as power consumption, environmental interference, static stability, etc. At the end of the process a numerical value is given to the action plan. The action plan that is good enough to enable the robot to accomplish its mission with the least possible consumption of power is given a higher fitness value. The simulation must be kept as simple as possible because it has to be run every single time an action plan is to be evaluated [49]. Also, there must be accuracy in the representation of the rover and its environment since the limitations of this approach are none but is depending on the accuracy of the model [61]. An accomplishment of challenging mission is only possible by utilizing more sophisticated operational methodologies, commanding scheme, and by means of proper data representation [62].

There are a couple of factors that permit the genetic algorithm search to find a possible solution. One of them is the way the fitness function assigns a numerical value to an action plan [49, 55]. During the evaluation of a certain plan, the simulation implements the plan as long as a physical constraint is not violated or the target is not reached [63]. The fitness of the plan is determined by a successful beginning portion. The plan that moves the robot a partial ahead towards the target receives a higher score as compared to a plan that is ineffective in doing the same job [49].

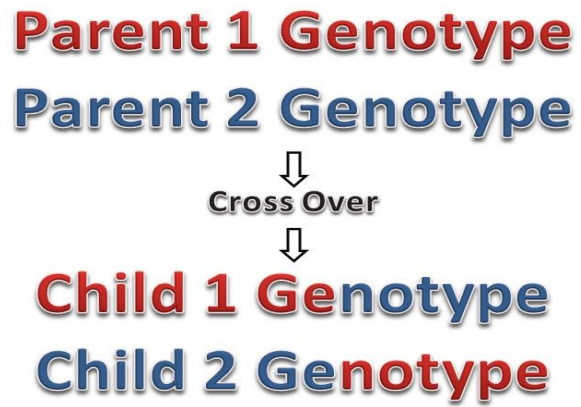

Fig. 10 - Working principle of a genetic crossover

The second contributing factor is the way the genetic crossover takes place. In the process, the combination of two plans of high fitness is carried out in order to produce two plans of even better fitness. However, the combination of both factors i.e. the fitness evaluation and the crossover method, allows one to build on action plans that are partially but not completely successful. Thus it implies that the designer is constructing on partly fruitful plans eyeing for means that lead to improvement rather than looking for a complete plan of action for task completion, and is instead building on partly positive plans by looking for methods to mend them [49].

## V. CONCLUSION

Space robots are the newest trend in space exploration. They not only offer a broader search aspect and freedom, but also go far beyond the physical limitations of humans. With a few planetary bodies already being explored and many exploratory missions in the pipeline, there is a need for optimal performing robots. Implementing optimization will offer countless benefits for autonomous space robots. In order to make sure the planetary robotic explorers accomplish their tasks and mission goals, certain methods are required to plan their action. With the availability of new information there is a need for developing new plans. The procedure of action plan generation aids in developing on develops a plan that enables there rover to accomplish the task without disrupting any of the physical constraints of the problem. This methodology thus helps in utilizing the capabilities of a robot and preventing the arrangement from being terminated. The modest examination of the methodology puts forward that it is more useful and applicable than a mathematical analysis may propose. Genetic algorithm plays an important role in finding ideal solutions to optimized space robots. A discussion of the process of selection, its nature and guiding principle for this process were presented after the demonstration of the procedure was carried out on a simple mobility task.

# VII. APPENDIX

## Table 1 - Motivation from previous research

| | Perception | Automation | Controls |
|---|---|---|---|
| **Automobile Robotics** | RC Arkin (1987), R Bajcsy (1988), DA Pomerleau (1992), AK Mackworth (1992), W Li (1994), H Kauppinen, T Seppanen (1995), A Arsenio, P Fitzpatrick (2003), C Pradalier, J Hermosillo, C Koike, C Braillon (2005), S Thrun, W Burgard, D Fox (2005). | J Barraquand, JC Latombe (1989), M Khatib, H Jaouni, R Chatila (1997), RM Murray, SS Sastry (1993), C Samson (1993), A De Luca, G Oriolo, C Samson (1998), P Švestka, MH Overmars (1998), H Chen, W Sheng, N Xi, M Song (2002), THS Li, SJ Chang (2003), C Pradalier, J Hermosillo, C Koike, C Braillon (2005), NB Hui, V Mahendar, DK Pratihar (2006). | KS Fu, RC Gonzalez, CSG Lee (1987), C Samson (1993), I Kolmanovsky, NH McClamroch (1995), C Samson (1995), A De Luca, G Oriolo, C Samson (1998), D Gu, H Hu (2002), TS Li, SJ Chang, YX Chen (2003), THS Li, SJ Chang (2003), CL Hwang, LJ Chang, YS Yu (2007), MW Spong, M Vidyasagar (2008). |
| **Marine Robotics** | G Russell, DM Lane (1986), A Elfes (1989), E Krotkov, J Blitch (1999), A Billard (2002), CE Lathan, M Tracey (2002), JG Bellingham, K Rajan (2007), D Ribas, N Palomeras, P Ridao (2007), R Bogue (2008), M Dunbabin, L Marques (2012), FS Hover, RM Eustice, A Kim, B Englot (2012). | TB Sheridan - Automatica (1989), JJ Leonard, HF Durrant (1992), JJ Leonard, AA Bennett, CM Smith, H Feder (1998), HJS Feder, JJ Leonard, CM Smith (1999), A Pascoal, P Oliveira, C Silvestre (2000), R Vaidyanathan, HJ Chiel, RD Quinn (2000), P Encarnaçao, A Pascoal (2001), D Soetanto, L Lapierre (2003), MR Benjamin, JA Curcio, JJ Leonard (2006), B Siciliano, O Khatib (2008). | J Yuh (1990), J Yuh (1995), A Pascoal, P Oliveira, C Silvestre (2000), J Yuh (2000), P Encarnaçao, A Pascoal (2001), L Lapierre, D Soetanto (2003), J Yu, M Tan, S Wang, E Chen (2004), L Lapierre, D Soetanto, A Pascoal (2004), MR Benjamin, JA Curcio, JJ Leonard (2006), TI Fossen (2011). |
| **Space Robotics** | Crowley, J. L. (1987), Kohn, W., & Skillman, T. (1988), DB Rubin (1988), Goldberg, D. (1989), A Elfes (1989), G Giralt, R Chatila, M Vaisset (1990), A Elfes (1991), A Elfes (1992), Ahuactzin, J.-M., Talbi, E.-G., Bessiere, P., & Mazer., E. (1992), DA Pomerleau (1992), G Hirzinger, J Heindl, K Landzettel (1992), RC Bolles (1992), P Moutarlier, R Chatila (1993), B Brunner, G Hirzinger, K Landzettel (1993), SB Kang, K Ikeuchi (1993), G Hirzinger, B Brunner, J Dietrich (1993), G Hirzinger, G Grunwald, B Brunner, J Heindl (1993), G Hirzinger, B Brunner, J Dietrich (1994), G Hirzinger (1994), V Santos, JGM Gonçalves, F Vaz (1994), Ram, A. G. (1994), Schultz, A. C. (1994), Schütte, N., Kelleher, J., & Namee, B. M. (1994), Toogood, R., Hao, H., & Wong, C. (1995), JD Nicoud, MK Habib (1995), E Kruse, R Gutsche, FM Wahl (1996), JL Crowley (1996), A Elfes (1996), G Oriolo, G Ulivi, M Vendittelli (1997), SB Kang, K Ikeuchi (1997), NI Katevas, NM Sgouros, SG Tzafestas (1997), Y Yamada, Y Hirasawa, S Huang (1997), Farritor, S., & Dubowsky, S. (1997), Yamauchi, B. (1997), Putz, | O Khatib (1987), G Hirzinger, B Brunner, J Dietrich (1994), YL Gu, Y Xu (1995), BL Ma, W Huo (1996), M Oda, K Kibe, F Yamagata (1996), M Oda (1999), CS Lovchik, MA Diftler (1999), K Yoshida, K Hashizume2001), C Li (2002), WK Yoon, T Goshozono, H Kawabe (2004). | GN Saridis (1983), MW Spong, M Vidyasagar (1987), Y Umetani, K Yoshida (1989), E Papadopoulos, S Dubowsky (1991), E Papadopoulos, S Dubowsky (1991), S Dubowsky, E Papadopoulos (1993), MW Spong (1995), KH Tan, MA Lewis (1996), C Sallaberger, SPT Force, CS Agency (1997), O Khatib (1999). |

| | P. (1998), S Thrun, D Fox, W Burgard (1998), W Burgard, AB Cremers, D Fox, D Hähnel (1998), C Weisbin, J Blitch, D Lavery, E Krotkov, C Shoemaker (1998), Cernic, S., Jezierski, E., Britos, P., Rossi, B., & Martínez, R. G. (1999), Ehlert, P. A. (1999), R Simmons, D Apfelbaum, W Burgard, D Fox, M Moors (2000), D Fox, W Burgard, H Kruppa, S Thrun (2000), D Fox, S Thrun, W Burgard, F Dellaert (2001), S Thrun, D Fox, W Burgard, F Dellaert (2001), A Howard, E Tunstel, D Edwards (2001), Thorpe, C., Clatz, O., Duggins, D., Gowdy, J., MacLachlan, R., Miller, J. R., et al. (2001), Lewis, M. A. (2002), Macchelli, A., C. Melchiorri, R. C., & Guidetti., M. (2002), Messom, C. (2002), B Åstrand, AJ Baerveldt (2002), R Kikuuwe, T Yoshikawa (2002), CF DiSalvo, F Gemperle, J Forlizzi (2002), Y Kuroki, M Fujita, T Ishida, K Nagasaka (2003), Nicolescu, M. N., & Mataric., M. J. (2003), Pedersen, L., Kortenkamp, D., Wettergreen, D., & Nourbakhsh, I. (2003), Elshamli, A., Abdullah, H. A., & Areibi., S. (2004), Jenkins, O. C., Nicolescu, M. N., & Mataric, M. J. (2004), M Montemerlo, S Thrun (2004), Zhou, Y. (2005), JS Franco, E Boyer (2005), E Bayro-Corrochano (2005), M Blow, K Dautenhahn, A Apple (2006), Blow, M., Dautenhahn, K., Appleby, A., Nehaniv, C. L., & Lee, D. C. (2006), Ciftcioglu, Ö., Bittermann, M. S., & Sariyildiz, I. S. (2006), Sánchez, P. D. (2006), Bittermann, M. S., Sariyildiz, I. S., & Ciftcioglu, Ö. (2007), Menon, C., Broschart, M., & Lan, N. (2007), G Hoffman, C Breazeal (2007), K Dautenhahn (2007), Ismail, A.-T., Sheta, A., & Al-Weshah, M. (2008), Kazem, B. I., Mahdi, A. I., & Oudah, A. T. (2008), Qureshi, F., & Terzopoulos, D. (2008), Enedah, C., Waldron, K. J., & Gladstone., H. B. (2008), Nakhaei, A. (2009), Pivtoraiko, M., Nesnas, I. A., & Kelly, A. (2009), Bruemmer, D., Few, D., Kinoshita, R., & Kapoor, C. (2009), RB Rusu, IA Sucan, B Gerkey, S Chitta (2009), BD Argall, S Chernova, M Veloso (2009), Shamsinejad, P., Saraee, M., & Sheikholeslam, F. (2010), Knudson, M., & Tumer, K. (2011), Huwedi, A. S., & Budabbus, S. M. (2012), Krasny, D. P. (2012), Purian, F. K., | | |
|---|---|---|---|

| | Farokhi, F., & Nadooshan, R. S. (2012), Tsuda, S., & Kobayashi., T. (2012), A Elfes (2013), Oh, J., Suppe, A., Stentz, A., & Hebert, M. (2013), Steder, B. (2013), Y Yoshihara, D Tang, N Kubota (2013), N Onkarappa (2013), DE Sonnleithner (2013), L Rozo (2013), JL Wright, JY Chen, SA Quinn, MJ Barnes (2013), Mohammad, Y. F., & Nishida, T. (2014), Schilling, P. D. (2014). | | |
|---|---|---|---|